\documentclass{svproc}

\usepackage{graphicx}        % For including images
\usepackage{tikz}            % For TikZ diagrams
\usetikzlibrary{shapes.geometric, arrows, positioning}
\usepackage{amsmath}         % For mathematical equations
\usepackage{hyperref}        % For hyperlinks
\usepackage{geometry}        % For adjusting margins
\usepackage{newtxtext,newtxmath}
\usepackage{microtype}
\usepackage{url}
\usepackage[T1]{fontenc}
\usepackage[utf8]{inputenc}
\usepackage{lmodern}
\usepackage{xcolor}
\usepackage{colortbl}
\usepackage{booktabs}
\usepackage{multirow}
\usepackage{authblk}
\usepackage{float}
\usepackage{adjustbox}       % For flexible figure placement
\title{Cross-Language Approach for Quranic QA}
\titlerunning{Quran QA}
\authorrunning{Islam Oshallah et al.}

\author{
    Islam Oshallah \and 
    Mohamed Basem \and 
    Ali Hamdi \and
    Ammar Mohammed 
}
\institute{
Faculty of Computer Science, MSA University, Egypt \\
\email{\{ islam.abdulhakeem, mohamed.basem1, ahamdi, ammohammed\}@msa.edu.eg}
}

\date{\today}
\begin{document}

\maketitle

\begin{abstract}
Question answering systems face critical limitations in languages with limited resources and scarce data, making the development of robust models especially challenging. The Quranic QA system holds significant importance as it facilitates a deeper understanding of the Quran, a Holy text for over a billion people worldwide. However, these systems face unique challenges, including the linguistic disparity between questions written in Modern Standard Arabic and answers found in Quranic verses written in Classical Arabic, and the small size of existing datasets, which further restricts model performance. To address these challenges, we adopt a cross-language approach by (1) {Dataset Augmentation}: expanding and enriching the dataset through machine translation to convert Arabic questions into English, paraphrasing questions to create linguistic diversity, and retrieving answers from an English translation of the Quran to align with multilingual training requirements; and (2) {Language Model Fine-Tuning}: utilizing pre-trained models such as BERT-Medium, RoBERTa-Base, DeBERTa-v3-Base, ELECTRA-Large, Flan-T5, Bloom, and Falcon to address the specific requirements of Quranic QA. Experimental results demonstrate that this cross-language approach significantly improves model performance, with RoBERTa-Base achieving the highest MAP@10 (0.34) and MRR (0.52), while DeBERTa-v3-Base excels in Recall@10 (0.50) and Precision@10 (0.24). These findings underscore the effectiveness of cross-language strategies in overcoming linguistic barriers and advancing Quranic QA systems.
\end{abstract}

\section{Introduction}

The Holy Quran is revered by Muslims worldwide as a Holy source of guidance and knowledge \cite{ali2000holy}. However, its accessibility is often limited by linguistic barriers, particularly for non-Arabic speakers and even for native Arabic speakers unfamiliar with its Classical Arabic form \cite{almelhes2024enhancing}. Readers and commentators have historically faced challenges in understanding the Quran due to its linguistic style and complex composition \cite{saeed2006interpreting}. These complexities are compounded by the linguistic gap between modern standard Arabic (MSA), often used in questions, and classic Arabic (CA), the language of the Quran \cite{kadhim2023translatability}. To bridge this gap, a cross-language approach is essential, taking advantage of advanced translations and language models.

Existing Quranic QA systems, such as those explored in the Quran QA 2023 shared task, have made significant strides to address these challenges \cite{malhas2023quranqa}. These tasks focus on two primary subtasks: Passage Retrieval, which aims to identify the most relevant Quranic passages in response to a query, and Machine Reading Comprehension, which aims to retrieve accurate and contextually relevant passages. However, the development of Quranic QA systems faces critical challenges due to the small size of existing datasets and the linguistic disparity between MSA, used in questions, and Classical Arabic (CA), found in Quranic verses. These limitations cause a severe hinder to the model performance, emphasizing the need for innovative approaches to dataset expansion and fine-tuning LLMs and LMs to meet the unique requirements of Quranic QA.

This research adopts a cross-language approach to enhance Quranic QA systems. By translating both questions and answers, the linguistic barriers posed by MSA and CA are overcome. The dataset was significantly expanded through machine translation, paraphrasing, and the integration of additional questions from existing literature. Fine-tuned LLMs, such as BERT-Medium, RoBERTa-Base, DeBERTa-v3-Base, ELECTRA-Large, Flan-T5, Bloom, and Falcon, were trained to address these challenges effectively. This approach not only enhances the dataset but also aligns advanced language models with the unique linguistic and contextual complexities of Quranic texts. By upgrading and highlighting these advancements, the system provides more accurate and context-sensitive responses, allowing deeper engagement with the Quran\cite{essam2024survey}.

Figure \ref{fig:workflow_rotated} illustrates the workflow adopted in this study, showcasing the process from dataset augmentation to fine-tuning and evaluation. This diagram highlights the cross-language approach and the transformations of the data set essential for improving the Quranic QA system.
\tikzstyle{data} = [rectangle, rounded corners, minimum width=1cm, minimum height=0.9cm, text centered, draw=black, fill=yellow!30, font=\scriptsize]
\tikzstyle{process} = [rectangle, minimum width=1.2cm, minimum height=0.6cm, text centered, draw=black, fill=blue!30, font=\scriptsize]
\tikzstyle{arrow} = [thick,->,>=stealth]

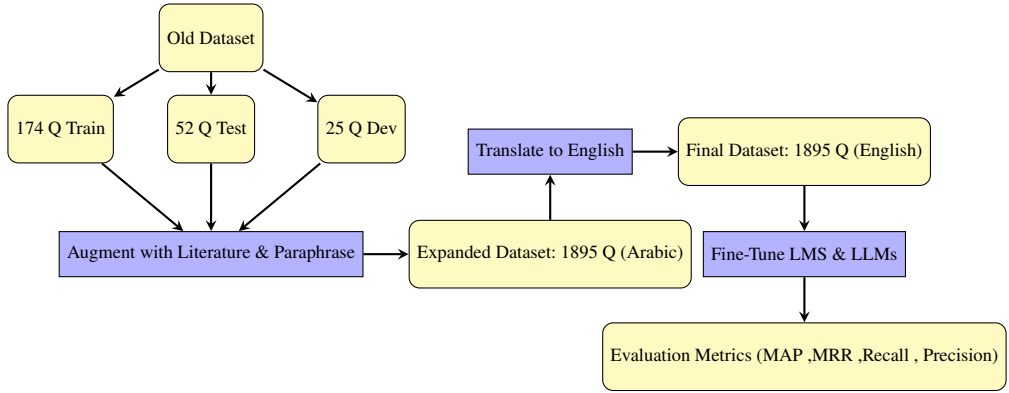
\begin{figure}[ht]
        \centering
        \begin{tikzpicture}[node distance=0.4cm and 0.6cm]
            % Old Dataset
            \node (OldDataset) [data] {Old Dataset};
            \node (TrainData) [data, below=0.3cm of OldDataset, xshift=-2cm] {174 Q Train};
            \node (TestData) [data, below=0.3cm of OldDataset] {52 Q Test};
            \node (DevData) [data, below=0.3cm of OldDataset, xshift=2cm] {25 Q Dev};
            % Dataset Expansion
            \node (AugmentLiterature) [process, below=0.9cm of TestData, text centered] {Augment with Literature \& Paraphrase};
            % New Dataset
            \node (ExpandedDataset) [data, right=0.6cm of AugmentLiterature, text centered] {Expanded Dataset: 1895 Q (Arabic)};
            % Translation
            \node (TranslateToEnglish) [process, above=0.6cm of ExpandedDataset, text centered] {Translate to English};
            % English Dataset
            \node (FinalDataset) [data, right=0.6cm of TranslateToEnglish, text centered] {Final Dataset: 1895 Q (English)};
            % Fine-Tuning
            \node (FineTuneLLMs) [process, below=0.6cm of FinalDataset, text centered] {Fine-Tune LMS \& LLMs};
            % Evaluation Metrics
            \node (EvaluationMetrics) [data, below=0.6cm of FineTuneLLMs, text centered] {Evaluation Metrics (MAP ,MRR ,Recall , Precision)};
            % Arrows
            \draw [arrow] (OldDataset) -- (TrainData);
            \draw [arrow] (OldDataset) -- (TestData);
            \draw [arrow] (OldDataset) -- (DevData);
            \draw [arrow] (TrainData) -- (AugmentLiterature);
            \draw [arrow] (TestData) -- (AugmentLiterature);
            \draw [arrow] (DevData) -- (AugmentLiterature);
            \draw [arrow] (AugmentLiterature) -- (ExpandedDataset);
            \draw [arrow] (ExpandedDataset) -- (TranslateToEnglish);
            \draw [arrow] (TranslateToEnglish) -- (FinalDataset);
            \draw [arrow] (FinalDataset) -- (FineTuneLLMs);
            \draw [arrow] (FineTuneLLMs) -- (EvaluationMetrics);
        \end{tikzpicture}
        \caption{\scriptsize Workflow for Cross-Language Dataset Expansion and Fine-Tuning.}
        \label{fig:workflow_rotated}

\end{figure}
\section{Related Work}
The Quranic text is rich and contains many events, making it essential to efficiently retrieve passages and answer questions \cite{qamar2024benchmark}. Its unique features, like complex syntax, multiple definitions, and deep linguistic elements, present challenges for question-answering systems \cite{essam2024decoding}. One major challenge is bridging the gap between Modern Standard Arabic and Classical Arabic, as well as handling unanswerable queries using techniques like thresholding \cite{elkomy2023tce}.

Models like AraBERT, CAMeLBERT, and AraELECTRA have shown great potential in handling Arabic text, effectively capturing its meaning and context \cite{aljamel2024fine}. However, limited or unbalanced datasets often hinder their performance, making it harder to handle new or unseen queries \cite{liu2023imbalanced}. Sarhan and Elkomy \cite{elkomy2023tce} addressed these issues by using hybrid models that combine dual-encoder and cross-encoder approaches. They leveraged transfer learning with external datasets like TyDI QA and Tafseer, improving performance with a MAP score of 0.25. Despite this, there is still a need for more dataset expansion.

Mahmoudi et al. \cite{mahmoudi2023multi} introduced a multitask transfer learning method, combining both unsupervised and supervised fine-tuning using models like AraELECTRA and AraBERT. They used advanced embedding techniques like TSDAE and SimCSE to improve passage retrieval by producing high-quality sentences. However, they also stressed the need for more dataset expansion.

In another approach, Alawwad et al. \cite{alawwad2023ahjl} focused on Quranic passage retrieval by translating Arabic queries into English. This allowed them to use advanced English-based models, such as OpenAI embeddings and sentence transformers. Their system also included a paraphrasing module to create multiple query variations, significantly improving retrieval accuracy. This method demonstrated the power of translation-based techniques in overcoming linguistic barriers. The queries were applied to an English-translated version of the Quran, utilizing LLMs and other models that support English \cite{pavlova2023leveraging}.

This work approach focuses on two main aspects: (1) Dataset Augmentation, which expands and enriches datasets by translating Arabic questions into English and paraphrasing to create more diverse queries. Answers are then retrieved from an English translation of the Quran; and (2) Language Model Fine-Tuning, where models like BERT-Medium, RoBERTa-Base, DeBERTa-v3-Base, ELECTRA-Large, Flan-T5, Bloom, and Falcon are fine-tuned to handle Quranic QA tasks. This approach effectively bridges the gap between Modern Standard Arabic and Classical Arabic, improving performance.

\section{Methodology}
The methodology of this study comprises three key phases:

\begin{enumerate}
\item \textbf{Dataset Preparation:} Expanding and diversifying the original dataset through the addition of new questions and paraphrasing to increase the linguistic variation.
\item \textbf{Cross-Language :} Translating questions into English to leverage the capabilities of widely available pre-trained language models that perform better in English.
\item \textbf{Model Fine-Tuning:} {\small Fine-tuning state-of-the-art models, including BERT-Medium, RoBERTa-Base, DeBERTa-v3-Base, ELECTRA-Large, Flan-T5, Bloom, and Falcon, to optimize performance for the Quranic QA task.}

\end{enumerate}

\subsection{Dataset Preparation}

\subsubsection{Database Collection}
The dataset was collected and picked from multiple authentic and trustworthy sources to ensure its diversity, accuracy, and relevance. This comprehensive dataset enhances the ability of the model to handle various QA scenarios and context-dependent or ambiguous queries from the Quranic text, thereby improving the overall performance of Quranic QA systems. The primary sources included:

\begin{itemize}
\item \textbf{Quran QA 2023 Shared Task Dataset:} This foundational dataset, curated for the Quran QA shared task \cite{malhas2023quranqa}, comprises 251 structured questions paired with annotated Quranic passages. It served as the basis for our experiments.
\item \textbf{SQuAD v2 Dataset:} The SQuAD v2 dataset \cite{squad_v2} was incorporated to increase diversity and address unanswerable questions. With over 100,000 questions, it enriched the adaptability of the model across various QA scenarios.
\item \textbf{1000 Questions and Answers in the Holy Quran PDF:} Extracted data from this resource \cite{ashor2023noor} underwent rigorous cleaning to ensure relevance and integration into the dataset.
\item \textbf{List of Plants Citation in Quran and Hadith:} Context-specific references from the Quranic Botanic Garden \cite{plants2024citation} added unique dimensions to the dataset.
\item \textbf{Translated Quran Texts:} Translations of the Quran \cite{quran_en_translations} bridged linguistic and contextual gaps between Modern Standard Arabic and Classical Arabic.
\end{itemize}

\subsubsection{Dataset Expansion}
The dataset was expanded from the original 251 questions in the Quran QA 2023 Shared Task to 629 questions using resources such as the \textit{1000 Questions and Answers in the Holy Quran} \cite{ashor2023noor} and the \textit{List of Plants Citation in Quran and Hadith} \cite{plants2024citation}. Each question was rephrased twice, resulting in a rich dataset of 1,895 questions categorized into single-answer, multi-answer, and zero-answer types.

This process, illustrated in Figure~\ref{fig:workflow_rotated}, demonstrates how dataset manipulation enhances linguistic diversity and adaptability. The expanded dataset allowed for fine-tuning multiple pre-trained transformer models, improving their ability to handle varied question formats and enhancing overall retrieval performance in Quranic QA tasks.

\subsubsection{Dataset Cleaning}
To ensure the quality of the dataset and reliability, the following cleaning steps were performed:
\begin{itemize}
\item \textbf{Removal of Irrelevant or Low-Quality Pairs:} Unclear or irrelevant question-passage pairs were removed to maintain dataset integrity.
\item \textbf{Standardization of Question Formats:} Questions were standardized to ensure consistency across the dataset, facilitating effective model training and comprehension.
\end{itemize}

\subsection{Cross-Language Approach}

A key aspect of this study is the cross-language approach, which involved translating the dataset into English to enhance the performance of language models in the Quranic Question Answering task. This approach addresses the linguistic challenges posed by the use of Modern Standard Arabic (MSA) in the questions and Classical Arabic (CA) in the Quranic passages. Since many pre-trained large language models (LLMs) and language models (LMs), such as BERT-Medium, RoBERTa-Base, DeBERTa-v3-Base, ELECTRA-Large, Flan-T5, Bloom, and Falcon, are primarily optimized for English, this translation step was essential to bridge the linguistic gap and significantly improve model performance in handling context-dependent or ambiguous queries from the Quranic text.

To make the dataset compatible with these models, the questions were translated using the Google Translate API To English, with a focus on preserving the structure and nuances of the original text. This process helped align the dataset with the strengths of pre-trained LMS and LLMs, which tend to perform more effectively on English data \cite{ali2023large}. Additionally, the use of the translated Quran by Marmaduke Pickthall \cite{quran_en_translations}, a renowned Islamic scholar, helped overcome the linguistic complexities associated with Classical Arabic. This translation, known for its authenticity and clarity, provided a standardized and contextually accurate version of the Quranic passages.

This translation process proved instrumental in improving the ability of the model to interpret user queries and retrieve relevant Quranic passages. By converting questions into English, the models could better understand and process the input while providing answers rooted in accurate and clear translation of Pickthall. This ensured that the answers preserved the original meanings while seamlessly adapting to the English language structure, making them both accessible and contextually accurate.

Furthermore, this cross-language approach helped address several challenges unique to Arabic, such as its complex syntax and linguistics, which can sometimes hinder the effectiveness of Arabic-language models. Translating both the questions and the Quranic verses into English facilitated the training of models that could generalize across different question formats and languages, improving overall performance.

In summary, the cross-language approach, which involved translating the dataset using the Google Translate API and utilizing the translated Quran of Marmaduke Pickthall, significantly enhanced the ability of the model to perform in Quranic QA tasks. It ensured the compatibility of state-of-the-art models with the dataset while preserving the integrity of the Quranic text, leading to improved retrieval accuracy and more context-aware answers.

\begin{figure}
    \centering
    \includegraphics[width=0.44\linewidth]{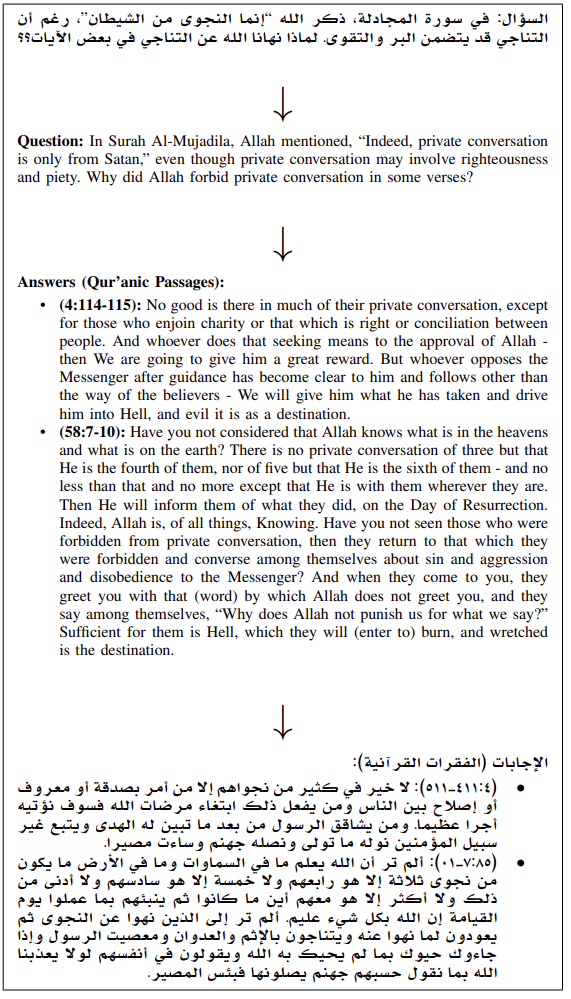}
 \caption{Cross-Language Example of Quranic QA: A question posed in Arabic is translated into English by using Google Translate API,then used to retrieve relevant passages from the English-translated Quran, and then translated back into Arabic. }
    \label{fig:enter-label}
\end{figure}

\subsection{Model Fine-Tuning}
In this study, transformer-based language models were fine-tuned for the Quran QA task, including BERT-Medium, RoBERTa-Base, DeBERTa-v3-Base, ELECTRA-Large, Flan-T5, Bloom, and Falcon. These models were pre-trained and fine-tuned on the SQuAD v2 dataset \cite{squad_v2} to improve their ability to handle complex and zero-answer questions. Each model brings unique strengths, collectively offering a robust solution for question answering. Additionally, a cross-encoder model was implemented in the study to enhance the way questions and texts interact. The algorithm was able to increase accuracy and better convey context by evaluating them as a single input, particularly when it came to recognizing the most significant portions. This methodology significantly contributed to delivering accurate responses that were also contextually appropriate.
\begin{table}[ht]
    \centering
    \caption{Summary of Models Fine-Tuned on SQuAD v2 Dataset (all models fine-tuned on SQuAD v2)}
    \begin{tabular}{|p{4.5cm}|c|p{5.5cm}|}

        \hline
        \textbf{Model}          & \textbf{Number of Parameters} & \textbf{Overview} \\ 
        \hline
        BERT-Medium             & 86 million   & Compact transformer designed for performance and efficiency \cite{devlin2019bert} \\ 
        \hline
        RoBERTa-Base            & 125 million  & Optimized for natural language understanding tasks \cite{liu2019roberta} \\ 
        \hline
        DeBERTa-v3-Base         & 220 million  & Enhances BERT and RoBERTa with disentangled attention \cite{he2020deberta} \\ 
        \hline
        ELECTRA-Large           & 335 million  
        & Efficient training method using a generator-discriminator architecture \cite{clark2020electra} \\ 
        \hline
        Bloom                   & 7 billion    & Large transformer model supporting a wide range of languages \cite{workshop2022bloom} \\ 
        \hline
        Falcon                  & 7 billion    & Designed for complex, context-dependent query handling \cite{almazrouei2023falcon} \\ 
        \hline
        Flan-T5                 & 11 billion   & Instruction-tuned for a variety of NLP tasks, including translation and question answering \cite{raffel2020exploring}\\ 
        \hline
    \end{tabular}
    \label{tab:models_summary}
\end{table}
\tikzstyle{process} = [rectangle, rounded corners, minimum width=2.5cm, minimum height=1cm, text centered, draw=black, fill=blue!20, font=\small]
\tikzstyle{data} = [rectangle, minimum width=2.5cm, minimum height=1cm, text centered, draw=black, fill=red!20, font=\small]
\tikzstyle{circlebox} = [circle, minimum width=1cm, minimum height=1cm, text centered, draw=black, fill=green!20, font=\small]
\tikzstyle{arrow} = [thick, ->, >=stealth]

\begin{figure}[ht]
    \centering
    \begin{tikzpicture}[node distance=0.5cm, every node/.style={scale=0.8}] % Reduced spacing and scaled nodes
    % Q-Text
    \node (qText) [data, text centered] {Q-Text};
    % Positive documents
    \node (posDocsLabel) [text centered, below=0.4cm of qText, xshift=-1.5cm, font=\small] {Positive passages};
    \node (posDocs) [draw, rectangle, minimum width=2.5cm, minimum height=1.2cm, below=0.3cm of posDocsLabel, font=\small] {
        \begin{tabular}{l|c}
        \textbf{Text} & \textbf{Relevance} \\ \hline
        Passage1 & 1 \\
        Passage3 & 1 \\
        ... & 1
        \end{tabular}
    };
    % Negative documents
    \node (negDocsLabel) [text centered, right=1cm of posDocsLabel, font=\small] {Negative passages};
    \node (negDocs) [draw, rectangle, minimum width=2.5cm, minimum height=1.2cm, below=0.3cm of negDocsLabel, font=\small] {
        \begin{tabular}{l|c}
        \textbf{Text} & \textbf{Relevance} \\ \hline
        Passage6 & 0 \\
        Passage14 & 0 \\
        ... & 0
        \end{tabular}
    };
    % Randomizer
    \node (randomizer) [process, below=0.5cm of posDocs, xshift=1.5cm] {Randomizer};
    % Combined data
    \node (combinedLabel) [below=0.4cm of randomizer, font=\small] {Combined Data for each question};
    \node (combined) [draw, rectangle, minimum width=3cm, minimum height=1.5cm, below=0.3cm of combinedLabel, font=\small] {
        \begin{tabular}{l|c}
        \textbf{Text} & \textbf{Relevance} \\ \hline
        Passage3 & 1 \\
        Passage6 & 0 \\
        ... & ...
        \end{tabular}
    };
    % LM using Cross-Encoder
    \node (lm) [process, right =0.5cm of combined] {LLM \& LM using Cross-Encoder};
    % LM Output
    \node (embedding) [process, right=0.4cm of lm] {LLM \& LM Output};
    % Retrieval
    \node (retrieval) [process, above=0.8cm of embedding] {Retrieval};
    % Test Cases
    \node (testCases) [data, above=0.8cm of retrieval] {Test Cases};
    % Answers
    \node (answersLabel) [right=0.9cm of testCases, font=\small] {Answers};
    \node (answers) [draw, rectangle, minimum width=2.5cm, minimum height=1.2cm, below=0.3cm of answersLabel, font=\small] {
        \begin{tabular}{l}
        Passage2 \\
        Passage5 \\
        Passage9 \\
        ...
        \end{tabular}
    };
    % Arrows
    \draw [arrow] (qText) -- (posDocsLabel);
    \draw [arrow] (qText) -- (negDocsLabel);
    \draw [arrow] (posDocs) -- (randomizer);
    \draw [arrow] (negDocs) -- (randomizer);
    \draw [arrow] (randomizer) -- (combined);
    \draw [arrow] (combined) -- (lm);
    \draw [arrow] (lm) -- (embedding);
    \draw [arrow] (embedding) -- (retrieval);
    \draw [arrow] (retrieval) -- (answers);
    \draw [arrow] (testCases) -- (retrieval); % Arrow connecting testCases to retrieval

    \end{tikzpicture}
    \caption{Workflow for LLM \& LM training}
    \label{fig:lm_workflow}
\end{figure}
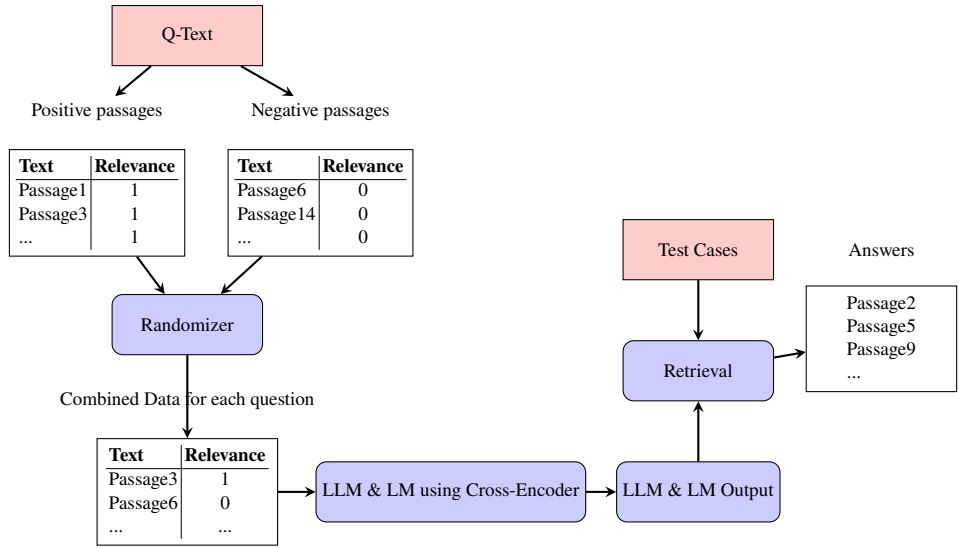

\subsection{Explanation of the Workflow Diagram}

Figure \ref{fig:lm_workflow} illustrates the complete workflow for fine-tuning and retrieval tasks in our Quran QA system. The diagram shows how the input question (Q-Text) is processed along with positive (relevant) and negative (non-relevant) passages to generate the final answer during the fine-tuning phase.

1. \textbf{Question-Text (Q-Text)}: In the training phase, the input question is provided as part of the dataset. This is a question from the dataset that the model is being trained to answer from the Quranic passages.

2. \textbf{Positive and Negative Passages}: The system uses passages from the Quran that are either relevant (positive passages) or irrelevant (negative passages) to the given question. These passages are labeled with "1" indicating relevance and "0" indicating irrelevance during training. This helps the model learn how to find relevant passages and handle context-dependent or ambiguous queries from the Quranic text.

3. \textbf{Randomizer and Combined Data}: The system combines both positive and negative passages using a randomizer . After mixing positive and negative passages with a randomizer, contrastive learning could further help the model by creating training pairs. Each pair would be evaluated to ensure that the model distinguishes between relevant and irrelevant passages effectively. This would enhance the ability of the model to prioritize relevant passages over irrelevant ones, improving retrieval performance.

4. \textbf{Cross-Encoder Fine-Tuning}: The combined data, along with relevant labels, is passed into a language model and large language model using a cross-encoder architecture to understand the context-dependent or ambiguous queries from the Quranic text. Models like RoBERTa-Base, DeBERTa-v3-Base, ELECTRA-Large, Flan-T5, Bloom, Falcon, and BERT-Medium have been fine-tuned using the Quran QA dataset. While these models are fine-tuned on other datasets like squad, then undergo specific fine-tuning for this task to better retrieve the most relevant passages when answering questions.

5. \textbf{LLM \& LM Output and Retrieval}: The fine-tuned model generates output in the form of relevance scores for each passage based on its understanding of the relationship between the input question and the passages.These scores are then used to rank the passages, with the highest ranked ones selected to provide the best and most relevant answers which led to an increase in the ability to deal with context-dependent or ambiguous queries from the Quranic text.

\section{Results}

The evaluation results show the impact of dataset expansion, cross language, and model fine-tuning on enhancing the performance of QA systems for the Holy Quran. Multiple models evaluated on metrics such as \textbf{MAP10} (Mean Average Precision at 10), \textbf{MRR} (Mean Reciprocal Rank), \textbf{Recall} at top 5 and 10 passages (\textbf{Rec5}, \textbf{Rec10}), and \textbf{Precision} for the same ranks (\textbf{Pre5}, \textbf{Pre10}).
\begin{table}[ht]
\centering

\label{tab:All_Results}
\footnotesize
\setlength{\tabcolsep}{0.8pt}
\begin{tabular}{l|cc|cc|cc|cc|cc|cc}
\toprule
\textbf{Model} & \multicolumn{2}{c|}{\textbf{MAP10}} & \multicolumn{2}{c|}{\textbf{MRR}} & \multicolumn{2}{c|}{\textbf{Rec5}} & \multicolumn{2}{c|}{\textbf{Rec10}} & \multicolumn{2}{c|}{\textbf{Pre5}} & \multicolumn{2}{c}{\textbf{Pre10}} \\
\cmidrule(lr){2-3} \cmidrule(lr){4-5} \cmidrule(lr){6-7} \cmidrule(lr){8-9} \cmidrule(lr){10-11} \cmidrule(lr){12-13}
& \textbf{Base} & \textbf{Ours} & \textbf{Base} & \textbf{Ours} & \textbf{Base} & \textbf{Ours} & \textbf{Base} & \textbf{Ours} & \textbf{Base} & \textbf{Ours} & \textbf{Base} & \textbf{Ours} \\
\midrule
\textbf{Electra-Large} & 0.04 & \textbf{0.31} & 0.11 & \textbf{0.43} & 0.05 & \textbf{0.34} & 0.08 & \textbf{0.46} & 0.05 & \textbf{0.21} & 0.04 & \textbf{0.19} \\
\textbf{Bloom}         & 0.04 & \textbf{0.14} & 0.14 & \textbf{0.24} & 0.07 & \textbf{0.25} & 0.15 & \textbf{0.29} & 0.07 & \textbf{0.14} & 0.07 & \textbf{0.10} \\
\textbf{Flan-T5}       & 0.01 & \textbf{0.26} & 0.07 & \textbf{0.35} & 0.01 & \textbf{0.33} & 0.08 & \textbf{0.38} & 0.02 & \textbf{0.20} & 0.03 & \textbf{0.14} \\
\textbf{Falcon}        & 0.03 & \textbf{0.26} & 0.10 & \textbf{0.40} & 0.05 & \textbf{0.32} & 0.08 & \textbf{0.40} & 0.04 & \textbf{0.20} & 0.04 & \textbf{0.12} \\
\textbf{Roberta-Base}  & 0.10 & \textbf{0.34} & 0.22 & \textbf{0.52} & 0.19 & \textbf{0.36} & 0.25 & \textbf{0.43} & 0.13 & \textbf{0.22} & 0.12 & \textbf{0.20} \\
\textbf{Bert-Medium}   & 0.07 & \textbf{0.27} & 0.17 & \textbf{0.39} & 0.09 & \textbf{0.35} & 0.23 & \textbf{0.40} & 0.10 & \textbf{0.20} & 0.08 & \textbf{0.18} \\
\textbf{Deberta-v3-Base} & 0.08 & \textbf{0.32} & 0.12 & \textbf{0.47} & 0.10 & \textbf{0.46} & 0.15 & \textbf{0.50} & 0.09 & \textbf{0.25} & 0.08 & \textbf{0.24} \\
\bottomrule

\end{tabular}

\caption{Comparison of multiple model versions based on MAP10, MRR, Recall, and Precision evaluation metrics. The table highlights the improvements achieved by the fine-tuned models (Ours) compared to their baseline (Base) across the evaluation metrics.}
\end{table}
The table highlights the highest improvements achieved by fine-tuning . Noticed observations are:
\begin{itemize}
 \item \textbf{Electra-Large} achieved the highest improvements, with MAP10 improving from 0.04 to 0.31 and MRR from 0.11 to 0.43, Reflecting better ranking of relevant passages.
    \item \textbf{Roberta-Base} showed strong baseline performance and achieved further improvements after fine-tuning, with MAP@10 increasing from 0.10 to 0.34 and MRR enhancing from 0.22 to 0.52.
 
All models showed consistent improvements in Recall (\textbf{Rec5, Rec10}) and Precision (\textbf{Pre5, Pre10}), with DeBERTa-v3-Base achieving the best results, scoring 0.46 and 0.50 in Recall (\textbf{Rec5, Rec10}) and 0.25 and 0.24 in Precision (\textbf{Pre5, Pre10}), respectively.
\end{itemize}

This detailed comparison highlights the effectiveness of expanding the dataset,cross language and fine-tuning the models, showing how these improvements help QA systems better retrieve and rank Quranic passages.

\section{Conclusion}

This study addresses the unique challenges of Quranic QA systems by adopting a cross-language approach to reduce the linguistic difference between Modern Standard Arabic questions and Classical Arabic Quranic texts. By enhancing the dataset with techniques like machine translation, paraphrasing, and adding more varied question sets, and by fine-tuning both language models and large language models, this work achieved improvements in performance. Notably, RoBERTa-Base demonstrated the best MAP@10 (0.34) and MRR (0.52), while DeBERTa-v3-Base excelled in Recall@10 (0.50) and Precision@10 (0.24). These results underscore the effectiveness of cross-language strategies in overcoming linguistic barriers and enhancing Quranic QA systems. Future research will focus on further expanding dataset diversity, exploring alternative fine-tuning strategies, Fine-tuning multilingual models to improve accuracy and generalizability. This work lays a foundation for more accessible and robust Quranic QA systems, allowing users worldwide to engage more deeply with the Holy Quran.

\begin{flushleft}

\end{flushleft}

\end{document}